\documentclass[11pt, a4paper, logo, onecolumn,copyright]{hcp}

\usepackage[authoryear, sort&compress, round]{natbib}
\title{In-Situ Tweedie Discrete Diffusion Models}

\renewcommand{\today}{}

\usepackage[utf8]{inputenc}
\usepackage[T1]{fontenc}
\usepackage{hyperref}
\usepackage{url}
\usepackage{booktabs}
\usepackage{nicefrac}
\usepackage{microtype}
\usepackage{amsmath}
\usepackage{graphicx}
\usepackage{multicol}
\usepackage{siunitx}
\usepackage{array}
\usepackage[nameinlink]{cleveref}
\usepackage{bbm}
\usepackage{multirow}
\usepackage{subfig}
\usepackage{soul}
\usepackage{floatrow}
\usepackage{float}
\usepackage{wrapfig}
\usepackage{blindtext}
\usepackage{tablefootnote}
\usepackage{amsfonts}
\usepackage[flushleft]{threeparttable}
\usepackage{colortbl}
\usepackage{bbding}
\usepackage{lipsum}
\usepackage[ruled,vlined]{algorithm2e}

\graphicspath{{figure/}}

\usepackage{xspace}

\newfloatcommand{capbtabbox}{table}[][\FBwidth]

\newcommand{\draftonly}[1]{#1}
\newcommand{\eat}[1]{}
\renewcommand{\draftonly}[1]{}
\definecolor{darkgreen}{RGB}{0, 102, 0}

\crefformat{section}{\S#2#1#3}

\DeclareMathOperator*{\argmax}{arg\,max}

\makeatletter
\renewcommand{\thefootnote}{}%
\renewcommand{\@makefnmark}{}%
\renewcommand{\@makefntext}[1]{#1}%
\makeatother

\author{\textbf{\large{Xiao Li$^{1}$*, Jiaqi Zhang$^{1}$*, Shuxiang Zhang$^{1}$}, Tianshui Chen$^{3,4}$, Liang Lin$^{1,2,3}$, Guangrun Wang$^{1,2,3}$\dag}\\
$^1$Sun Yat-sen University, Guangzhou, China\\
$^2$Guangdong Key Laboratory of Big Data Analysis and Processing\\
$^3$X-Era AI Lab\\
$^4$Guangdong University of Technology\\
Emails: \texttt{\{lixiao68,zhangjq88,zhangshx36\}@mail2.sysu.edu.cn, wanggrun@gmail.com, linliang@ieee.org, chentianshui@gdut.edu.cn}\\
}

\begin{abstract}
While diffusion models excel at generating continuous data such as images, adapting them to discrete tasks has relied on indirect approaches that either operate in continuous embedding spaces or use token masking mechanisms, both of which deviate from modeling the true discrete data distribution that can be theoretically guaranteed by Tweedie’s formula. We propose \textbf{in-situ Tweedie Discrete Diffusion} (TDD), a framework that performs Tweedie's-formula–guaranteed diffusion directly within the discrete one-hot space—hence ``in-situ.'' Unlike prior methods that diffuse continuous embeddings or mask tokens, TDD directly corrupts one-hot vectors with Gaussian noise and performs iterative denoising through a timestep-conditioned cross-entropy objective rather than mean-squared-error reconstruction. At each denoising step, the model predicts class probabilities, applies argmax to obtain discrete predictions, converts them to one-hot vectors, and feeds them into the next iteration with progressively reduced noise. This process naturally unifies discriminative classification and generative modeling under a single framework. Experiments demonstrate that TDD achieves strong performance on both image classification and text generation tasks, with extensive ablation studies confirming the effectiveness of each design component. Our work establishes a principled approach to discrete diffusion that preserves the core characteristics of diffusion models while operating natively in discrete space.

\end{abstract}

\begin{document}

\maketitle

\renewcommand{\thefootnote}{\fnsymbol{footnote}} 
\footnotetext[1]{* Equal contribution.} 
\footnotetext[2]{\dag~Corresponding author.} 

\renewcommand{\floatpagefraction}{0.8}

\section{Introduction}
\label{sec:intro}

\begin{figure}[t]
	\centering
	\includegraphics[width=1.0\linewidth]{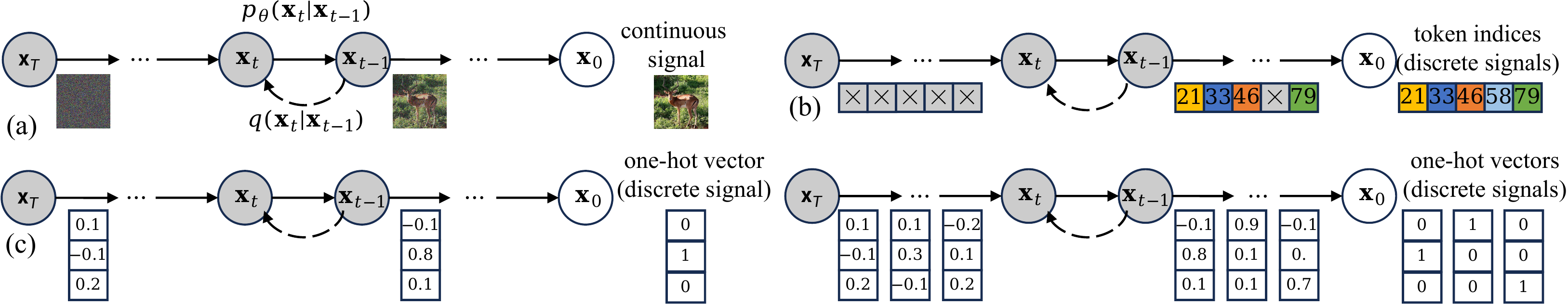}
	\caption{\textbf{Comparison between (a) traditional continuous-space diffusion, (b) Mask-based Discrete Diffusion (MDD), and (c) our proposed \emph{in-situ Tweedie Discrete Diffusion} (TDD) framework. In (c), the left panel illustrates single-token discrete generation (e.g., classification), while the right panel illustrates multi-token discrete generation (e.g., text generation).} Continuous diffusion models operate in Gaussian space, performing noise prediction and MSE-based reconstruction. Mask-based discrete models mimic diffusion through masked token recovery. In contrast, TDD begins from Gaussian-corrupted one-hot vectors and performs denoising directly in the one-hot space. At each step, the model applies an $\argmax$ to produce discrete predictions, converts them into one-hot vectors, and feeds them into the next iteration after adding noise with a reduced coefficient. This refinement process yields stable and efficient discrete-space diffusion, achieving accurate categorical predictions in only a few steps.
}
	\label{fig:front_page}
\end{figure}

Diffusion models have emerged as a powerful class of generative methods, achieving state-of-the-art performance in continuous domains by modeling data generation as a reverse denoising process~\citep{ho2020denoising}. These models operate in continuous-valued vector spaces and are typically trained using mean squared error (MSE) to reconstruct data from Gaussian noise. However, this formulation is inherently incompatible with discrete signals, whose categorical structure and non-Euclidean geometry introduce significant challenges for both optimization stability and sampling fidelity.

Efforts to extend diffusion models to discrete domains have generally followed two unsatisfactory paths. \textbf{The first line of research} formulates discrete diffusion as adding discrete noise, such as discrete Gaussian or discrete uniform perturbations~\citep{Austin2021StructuredDD,hoogeboom2021argmax,tran2019discrete}. Because these methods often yield poor performance, recent studies have shifted toward token masking strategies, where discrete noise is simulated by randomly replacing input tokens with a special mask token, and the model is trained to reconstruct the original sequence~\citep{gemini_diffusion2025,nie2025large,wu2025fast,dream2025}. However, this paradigm effectively replicates masked language modeling as in BERT~\citep{devlin2019bert} rather than performing true diffusion. All methods in this category—whether based on discrete Gaussian, uniform, or masking noise—fail to satisfy Tweedie's formula, and therefore lack the theoretical guarantee that the learned process faithfully models the true data distribution. As a result, their empirical performance typically remains inferior to strong autoregressive baselines.
\textbf{The second line of research} attempts to circumvent the discreteness issue by mapping tokens into continuous latent spaces via learnable embeddings~\citep{gong2022diffuseq}. While this approach enables the use of standard Gaussian diffusion techniques, it introduces new challenges: since the embedding space must itself be learned rather than fixed, injecting noise into it can lead to shortcut learning and unstable optimization dynamics. Consequently, generation quality and sample fidelity often degrade, especially when applied to complex discrete datasets.

In this work, we introduce the \textbf{in-situ Tweedie Discrete Diffusion (TDD)} model—a framework that preserves the defining properties of diffusion (i.e., Tweedie's formula) while operating directly in discrete spaces. TDD begins with Gaussian-corrupted one-hot vectors (i.e., ``in-situ'') and iteratively denoises them through multiple steps following a well-defined noise schedule. At each inference step, the model predicts a clean one-hot vector by applying an $\argmax$ operation to its output probabilities, converting the result into a one-hot representation, and re-injecting it into the next iteration with progressively reduced noise. This feedback loop enables progressive refinement, allowing the model to reduce uncertainty over time and converge toward semantically precise, categorical outputs.

Crucially, TDD employs a timestep-conditioned cross-entropy loss that directly enforces correspondence between predicted and target one-hot vectors, while mitigating over-reliance on conditioning signals during training. This design avoids the smoothing effects inherent in MSE losses and ensures that outputs follow the mutually exclusive nature of discrete categories. As illustrated in Fig.~\ref{fig:front_page}, our method differs fundamentally from prior approaches: continuous diffusion models operate in Gaussian space and perform regression-based noise prediction (Fig.~\ref{fig:front_page}a), while mask-based discrete models simulate diffusion through masked-token recovery (Fig.~\ref{fig:front_page}b). In contrast, TDD performs Tweedie’s-formula–guaranteed diffusion entirely within the one-hot space, using $\argmax$-based discretization and iterative noise reduction to progressively refine categorical predictions (Fig.~\ref{fig:front_page}c).

We evaluate TDD on two representative tasks, classification and text generation (such as image captioning), which serve as practical benchmarks for discrete generative modeling. The model demonstrates strong performance with only a few sampling steps, providing an efficient solution for generation in categorical domains.

Our key contributions are summarized as follows:
\begin{itemize}[label=\textbullet]
    \item We propose TDD, a discrete diffusion framework that preserves Tweedie's formulation while operating entirely within one-hot space-hence ``in-situ''.
    \item We introduce a timestep-conditioned cross-entropy loss that directly supervises categorical predictions, avoiding the smoothing effects of MSE and enforcing mutual exclusivity among discrete outputs.
    \item We design an iterative refinement mechanism that starts from Gaussian-corrupted one-hot vectors and progressively denoises them via $\argmax$-based discretization and reduced-noise feedback.
    \item We evaluate TDD on classification and text generation benchmarks, achieving strong performance with minimal sampling steps and providing extensive ablations to isolate component effects.
\end{itemize}
\section{Related Work}
\label{sec:related_work}

\paragraph{Diffusion Models in Continuous Space.}
Diffusion models have achieved state-of-the-art performance in continuous generative modeling, particularly in image synthesis. The foundational Denoising Diffusion Probabilistic Model (DDPM)~\citep{ho2020denoising} models data generation as the reverse of a Gaussian noising process, trained with a mean squared error (MSE) loss to predict added noise. Numerous extensions have enhanced efficiency and fidelity: DDIM~\citep{song2021denoising} accelerates generation via non-Markovian deterministic sampling; ADM~\citep{dhariwal2021diffusion} incorporates classifier guidance and adversarial techniques; and LDM~\citep{rombach2022highresolution} reduces computation through latent-space diffusion.
Despite their success, these methods inherently rely on continuous signal representations and quadratic losses, making them poorly suited for discrete data whose outputs lie on a simplex. One line of work attempts to address this by mapping discrete variables into continuous latent spaces via embedding~\citep{gong2022diffuseq}. However, such approaches often suffer from degraded generation quality and unstable training, as the continuous diffusion process struggles to model sharp, mutually exclusive distributions.

\paragraph{Masked-Discrete Diffusion Models.}
Extending diffusion models to discrete domains remains an open challenge. A common alternative reframes ``discrete diffusion'' as a masked token recovery task~\citep{gemini_diffusion2025,nie2025large,wu2025fast,dream2025}. In this view, noise is simulated through random masking of tokens, and the model is trained to reconstruct the original input, effectively mirroring masked language modeling as popularized by BERT~\citep{devlin2019bert}. While such models can be effective in certain language or multimodal applications, they deviate from the formal definition of diffusion~\citep{sohl2015deep}, lacking a principled forward noising process and stochastic transition dynamics.
Our proposed \textbf{TDD} framework addresses this conceptual gap by preserving the defining characteristics of diffusion in a truly discrete setting. Unlike embedding-based or masking-based approaches, TDD operates directly in one-hot space. In the \emph{forward} process, a clean one-hot vector is perturbed with Gaussian noise under a variance schedule, producing Gaussian-corrupted one-hot vectors. In the \emph{reverse} process, the model iteratively denoises these vectors: given a corrupted input, it predicts class probabilities, which are discretized via $\argmax$ into one-hot form. This estimate is supervised by timestep-conditioned cross-entropy alignment with the target one-hot vector, then re-noised with a reduced coefficient for the next iteration. This $\argmax$-and-re-noise loop constitutes a genuine diffusion mechanism over discrete symbols, supporting both single-token tasks (e.g., classification) and multi-token generation (e.g., text).

\paragraph{Bridging Generative and Discriminative Learning.}
There is growing interest in applying diffusion models or generative models to discriminative tasks such as classification \citep{wang2022traditional,wang2022semantic}, segmentation, and object detection. Existing methods often use diffusion indirectly—for example, as a feature extractor~\citep{zhu2024exploring}, a synthetic data generator~\citep{li2023open,nguyen2023dataset,ma2023diffusionseg,wu2023diffumask}, or a source of attention cues for zero-shot reasoning~\citep{yang2024exploring,liu2024vgdiffzero,ni2023ref}. Another line explores zero-shot generative classifiers, where classification is formulated as finding the class condition that minimizes diffusion loss~\citep{li2023your}. While promising, these approaches treat discriminative tasks as secondary outcomes of generative modeling, often requiring external guidance or downstream fine-tuning.
Recent works \citep{chen2025gs} also apply diffusion models to segmentation tasks and achieve remarkable performance, but these approaches still operate in continuous space and rely on MSE loss.
In contrast, TDD directly formulates discrete prediction as diffusion in one-hot space. For single-token tasks such as classification and multi-token tasks such as text generation, denoising itself serves as categorical prediction. This design eliminates the need for handcrafted supervision signals or embedding-space generation, realizing discrete generation through tweedie diffusion in symbolic space.

\begin{figure*}[t]
\vspace{-11pt}
	\centering
	\includegraphics[width=\linewidth]{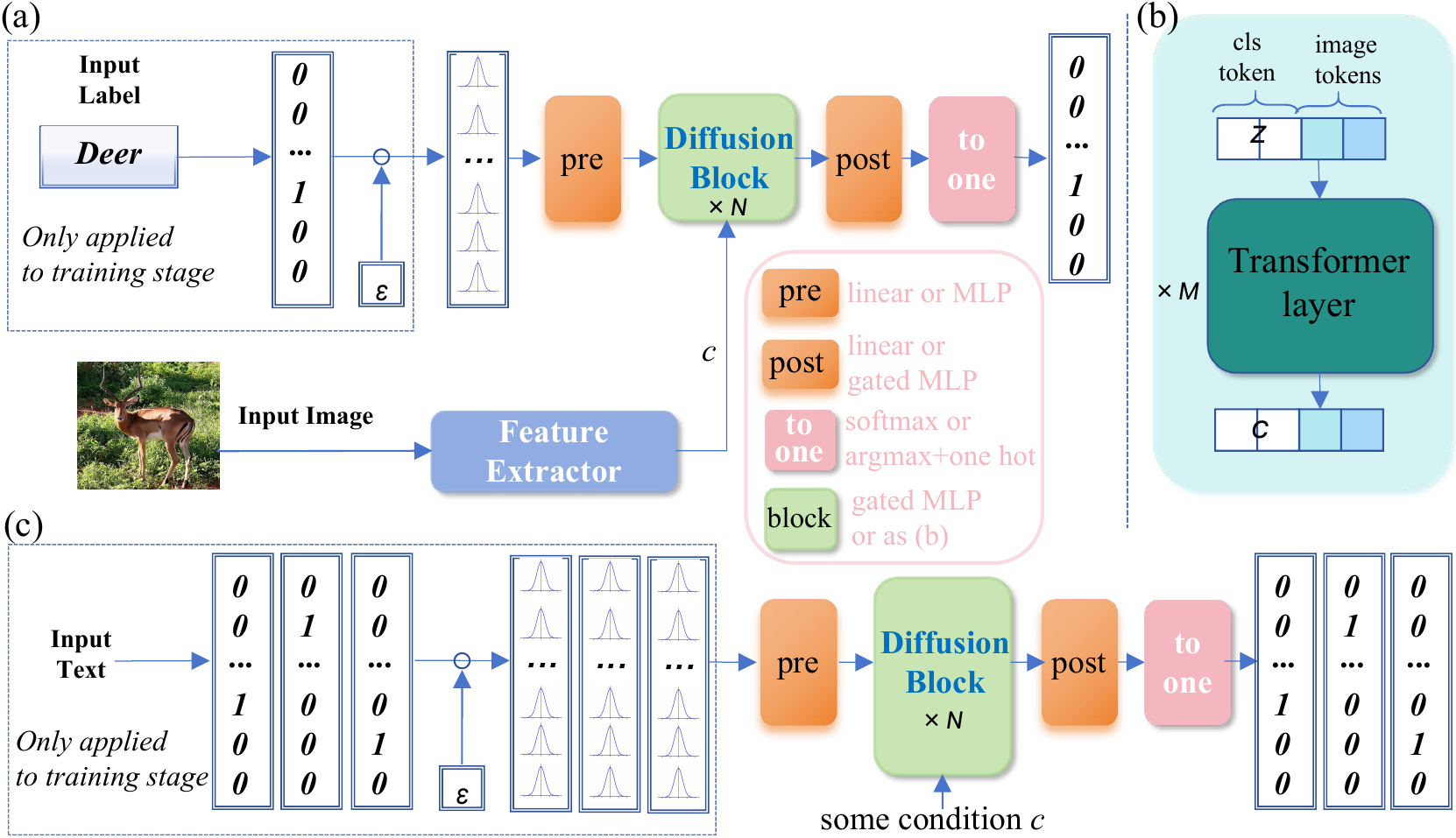}
    \vspace{-5pt}
	\caption{\textbf{Overview of the proposed framework.}
(a) \textit{Single-token generation for classification.} Ground-truth labels are converted into one-hot vectors and perturbed with Gaussian noise during training. The diffusion model iteratively denoises these corrupted vectors back to categorical one-hot outputs through the ``\textbf{to one}'' operation—implemented as softmax with timestep-conditioned cross-entropy supervision in training and as $\argmax$ with one-hot vectorization during sampling. Image features provide conditioning signals that guide the denoising process.
(b) \textit{Conditioning module.} The feature extractor encodes the input image into tokens, where some learnable class tokens interacts with image tokens through stacked Transformer layers to yield the conditioning representation $c$, which is injected into diffusion blocks.
(c) \textit{Multi-token generation for text.} TDD extends naturally from single-label classification to sequence generation by applying Gaussian corruption and denoising to each token in a sequence of one-hot vectors. The diffusion blocks predict categorical distributions for all tokens in parallel, enabling efficient iterative refinement of entire sequences into coherent text.
\vspace{-11pt}
}
	\label{fig:overview}
\end{figure*}
\section{Methodology}
\label{sec:method}

The goal of this work is to develop an \textbf{Tweedie Discrete Diffusion} (\textbf{TDD}) framework that retains the essential characteristics of diffusion models while operating directly in one-hot space. Unlike prior masked-discrete approaches that rely on masked-token recovery or diffusion in embedding space, TDD defines both forward and reverse processes directly on categorical one-hot vectors. 

The framework integrates three coordinated components: (i) a forward process that perturbs one-hot labels with Gaussian noise, (ii) a reverse process that iteratively refines predictions during inference, and (iii) a timestep-conditioned cross-entropy objective that supervises training. Conditioning signals (e.g., image features or other modalities) provide semantic guidance but do not replace the discrete generative process. An overview of the pipeline is shown in Fig.~\ref{fig:overview}.

\subsection{Background: Diffusion Models}

Diffusion models corrupt data progressively with Gaussian noise in a forward process and learn to recover the clean signal through a reverse process. Formally, given data $\mathbf{x}_0$, the forward distribution is:

\[
    q(\mathbf{x}_t | \mathbf{x}_0) = \mathcal{N}\left(\mathbf{x}_t; \sqrt{\bar{\alpha}_t}\mathbf{x}_0, (1-\bar{\alpha}_t)\mathbf{I}\right),
\]

where $\bar{\alpha}_t = \prod_{s=1}^t \alpha_s$ denotes the variance schedule. The reverse process is parameterized as:

\[
    p_{\theta}(\mathbf{x}_{t-1} | \mathbf{x}_t) = \mathcal{N}\big(\mathbf{x}_{t-1};\, \mu_\theta(\mathbf{x}_t, t), \sigma_t^2 \mathbf{I}\big).
\]

Standard diffusion models are trained by regressing Gaussian noise with an MSE objective, which is well-suited to continuous domains but fails to respect the mutually exclusive structure of categorical labels.

\subsection{Tweedie Discrete Diffusion}

TDD reformulates diffusion over categorical one-hot vectors. Let $\mathbf{y}_0 \in \{0,1\}^K$ denote a one-hot label over $K$ classes, with $\sum_{k=1}^K y_0^{(k)}=1$. We also experimented with $\lambda$-hot labels, where $\lambda \in (0,1)$, to make the representations more easily perturbed by Gaussian noise. While this variant led to lower training loss, it did not yield noticeable improvements in accuracy. For this reason, we omit detailed results here and leave further exploration of $\lambda$-hot labels to future work.

\paragraph{Forward process.}
At each step $t \in \{1,\dots,T\}$, the one-hot label is perturbed by Gaussian noise:

\[
    q(\mathbf{y}_t | \mathbf{y}_0) = \mathcal{N}\big(\mathbf{y}_t;\, \sqrt{\bar{\alpha}_t}\mathbf{y}_0,\,(1-\bar{\alpha}_t)\mathbf{I}\big),
\]

producing Gaussian-corrupted vectors $\mathbf{y}_t$ that remain close to the categorical simplex.

\paragraph{Reverse process (training).}
The denoising network parameterizes a categorical distribution:

\[
    p_\theta(\mathbf{y}_0 | \mathbf{y}_t, c) = \text{Softmax}\big(f_\theta(\mathbf{y}_t, t, c)\big),
\]

where $c$ denotes conditioning signals. Training minimizes the timestep-conditioned cross-entropy loss:

\[
    \mathcal{L}_{\text{CE}} = -\mathbb{E}_{t \sim \mathcal{U}[1,T]} \bar{\alpha}_t \sum_{k=1}^K y_0^{(k)} \log p_\theta(y_0^{(k)} | \mathbf{y}_t, c).
\]

The decay coefficient $\bar{\alpha}_t$ helps prevent the model from overly relying on the conditions of the conditional diffusion model, thereby promoting the effective learning of the diffusion network.

\paragraph{Reverse process (inference).}
At inference, predictions are discretized to remain in the one-hot space:

\[
    \hat{\mathbf{y}}_0 = \text{onehot}\Big(\argmax_k p_\theta(y_0^{(k)} | \mathbf{y}_t, c)\Big).
\]

The discretized prediction is re-noised with reduced variance:

\[
    \mathbf{y}_{t-1} \sim \mathcal{N}\left(\sqrt{\bar\alpha_{t-1}}\,\hat{\mathbf{y}}_0,\,(1-\bar\alpha_{t-1})\mathbf{I}\right).
\]

Iterating this ``argmax-and-re-noise'' loop progressively sharpens predictions until a clean one-hot label is recovered and we visualize this phenomenon in Fig~\ref{fig:heatmap}.

\subsection{Conditioning with Feature Extractors}

TDD incorporates conditional signals for guidance while preserving the discrete diffusion pipeline. For image-conditioned tasks, an encoder $\mathcal{E}$ maps an input $\mathbf{x}$ to tokens:

\[
    \mathbf{z} = \mathcal{E}(\mathbf{x}) \in \mathbb{R}^{L \times d}.
\]

A learnable class token $\mathbf{z}_{\text{cls}}$ is prepended, yielding

\[
    \tilde{\mathbf{z}} = [\mathbf{z}_{\text{cls}};\mathbf{z}_1,\dots,\mathbf{z}_L].
\]

After $M$ Transformer layers, either the updated class token or the average of other tokens serves as the conditioning vector:

\[
    c = \mathbf{z}_{\text{cls}}^{(M)}  \quad \text{or} \quad   c = \text{mean}(\mathbf{z}_{\text{1}}^{(M)}, \dots,\mathbf{z}_{\text{L}}^{(M)}).
\]

This vector $c$ is injected into the denoising network, as defined in Section~\ref{sect:task}.  
To reduce computation and prevent the learned condition from overshadowing the learning of the subsequent diffusion network, we adopt a feature reuse strategy inspired by \citep{li2024autoregressive}: (1) \textit{Expand} each feature batch $K$ times, (2) \textit{Sample} $K$ distinct timesteps per instance to generate noisy labels, and (3) \textit{Inject} the same features into all noisy labels. This $K$-fold strategy  enables multi-timestep optimization without repeated feature extraction.

\subsection{Single and Multiple Token Generation}
\label{sect:task}

\paragraph{Classification.}
For single-label classification, TDD diffuses and denoises one-hot vectors corresponding to class labels. Conditioning (e.g., image features) guides the denoising process, while timestep-conditioned cross-entropy supervision enforces categorical fidelity. We adopt a simple Transformer to inject conditioning into the diffusion block (see Fig.~\ref{fig:overview}). This setup demonstrates the stability and efficiency of TDD in the simplest categorical setting.

\paragraph{Text generation.}
TDD extends naturally to sequences. A sentence is represented as $\mathbf{Y}_0 = [\mathbf{y}_{0,1},\dots,\mathbf{y}_{0,N}]$, where each $\mathbf{y}_{0,i}$ is a one-hot vector over the vocabulary. The forward process perturbs each token independently:

\[
    q(\mathbf{y}_{t,i} | \mathbf{y}_{0,i}) = \mathcal{N}\big(\mathbf{y}_{t,i}; \sqrt{\bar{\alpha}_t}\mathbf{y}_{0,i}, (1-\bar{\alpha}_t)\mathbf{I}\big).
\]

The reverse process predicts distributions in parallel:

\[
    p_\theta(\mathbf{y}_{0,i} | \mathbf{y}_{t,i}, c) = \text{Softmax}\big(f_\theta(\mathbf{y}_{t,i}, t, c)\big).
\]

Training minimizes the sequence-level loss:

\[
    \mathcal{L}_{\text{text}} = -\sum_{i=1}^N \sum_{k=1}^K y_{0,i}^{(k)} \log p_\theta(y_{0,i}^{(k)} | \mathbf{y}_{t,i}, c).
\]

At inference, each token is discretized by $\argmax$, and iterative refinement sharpens the entire sequence into a coherent sentence. Unlike autoregressive models, TDD denoises all tokens simultaneously, enabling parallel and efficient text generation. Conditioning is injected via self-attention in the diffusion block (see Fig.~\ref{fig:overview}).
\section{Experiments}
\label{sec:experiment}

\subsection{Experimental Setup}

\paragraph{Datasets.} 
We evaluate TDD on two representative tasks: image classification and image-conditioned text generation. The former utilized the ImageNet dataset, while the latter employed both the imagenet-1k-vl-enriched\citep{imagenet-1k-vl-enriched} and COCO\citep{lin2014microsoft} datasets.  
For classification, we use the ImageNet benchmark \citep{5206848}, following the standard data splitting protocol. 
Based on experimental observations on the standard ImageNet, we conducted experiments on the imagenet-1k-vl-enriched\citep{imagenet-1k-vl-enriched} dataset to investigate the text generation performance of our method. However, given the limited prior work on this dataset, we transitioned to the COCO\citep{lin2014microsoft} to perform more comprehensive experiments, thereby ensuring a more robust and fair evaluation of TDD's performance.

\paragraph{Model and Training.} 
TDD integrates a Transformer feature extractor with a diffusion-based categorical denoiser, creating a hybrid architecture optimized for representation learning and generative refinement. Our full architecture contains \textbf{111 million} trainable parameters. The Transformer encoder dominates the parameter budget (\textbf{87M}) and specializes in hierarchical feature extraction, while the diffusion module contains \textbf{24M} parameters and is dedicated to discrete-space denoising. TDD is trained from scratch for \textbf{500 epochs} using AdamW \citep{Loshchilov2017DecoupledWD}. We adopt a base learning rate of $1\!\times\!10^{-4}$ with a weight decay of $0.3$ and effective batch size of $4096$, distributed across $8 \times$ NVIDIA A100 80GB GPUs. A warmup of $20$ epochs is followed by cosine learning rate decay \citep{Goyal2017AccurateLM}. Gradient clipping is applied with a global norm of $3.0$, and PyTorch AMP is used for mixed precision training. For captioning, we employed smaller batch sizes and weight decay values.

\paragraph{Competitor.} 
We compare our method against the strongest existing discrete diffusion baseline, the mask-based discrete diffusion model (MDD). We do not include other discrete diffusion variants such as D3PM~\cite{Austin2021StructuredDD}, ArgMax Flows~\cite{hoogeboom2021argmax}, multinomial diffusion~\cite{hoogeboom2021argmax}, or discrete flows~\cite{tran2019discrete}, as prior work has consistently shown that their performance is substantially weaker than that of MDD. MDD adopts the same Transformer backbone as our model. For multi-token prediction tasks (e.g., text generation), MDD follows the standard masked-token paradigm: it randomly masks a subset of tokens and predicts the masked locations. For single-token prediction tasks, since only one token is present, MDD replaces that token with a special mask symbol and predicts the ground-truth category. This is equivalent to appending a class token to the input and using its final-layer representation to perform single-token generation (i.e., classification).


\subsection{Single-Token Generation (Classification)}

\begin{table}[t]
\centering
\caption{\textbf{Comparison with state-of-the-art methods on ImageNet.} All models are trained in an end-to-end manner. Backbone architectures include ViT-Base/16, ViT-Large/16, and ViT-Huge/14~\citep{dosovitskiy2020image}. Results are reported at an input resolution of $224 \times 224$, with ViT-H additionally evaluated at $448 \times 448$. ``*'' denotes results reported by \citep{li2024autoregressive}. The best result is highlighted in \textbf{bold}.}
\label{tab:main}
\resizebox{0.9\linewidth}{!}{
\begin{tabular}{l|c|c|c|c}
    \toprule
    \textbf{Methods} & Resolution & Patch & \textbf{\#params} & \textbf{Top-1 (\%)} \\
    \toprule

    MDD-ViT-Base (standard classification) * 
        & $224 \times 224$ & $16 \times 16$ & $87\,\mathrm{M}$  & $82.3$ \\ 
    \midrule

    MDD-ViT-Large (standard classification) *
        & $224 \times 224$ & $16 \times 16$ & $305\,\mathrm{M}$ & $82.6$ \\ 
    \midrule

    MDD-ViT-Huge (standard classification) *
        & $448 \times 448$ & $14 \times 14$ & $632\,\mathrm{M}$ & $83.1$ \\ 
    \midrule

    \textbf{TDD-ViT-Base (w/ timestep-conditioned coefficients)} 
        & $224 \times 224$ & $16 \times 16$ & $111\,\mathrm{M}$ & $\mathbf{82.8}$ \\ 
    \midrule

    \textbf{TDD-ViT-Base (w/o timestep-conditioned coefficients)} 
        & $224 \times 224$ & $16 \times 16$ & $111\,\mathrm{M}$ & $\mathbf{83.0}$ \\ 
    \bottomrule
\end{tabular}
}
\end{table}

\subsubsection{Comparison with State-of-the-Art Methods}
We begin by comparing our approach with state-of-the-art models on ImageNet. As the state-of-the-art method, MDD uses the [CLS] token as the target of diffusion, with its fully uninformative form represented like MASK. It is designed to diffuse a single label from the MASK using the MDD approach. Our results are summarized in Tab.~\ref{tab:main}, from which two key observations emerge that highlight the advantages of our method.

\paragraph{TDD vs. MDD.} 
First, we consider a direct comparison against MDD. To ensure fairness, we construct a counterpart model by replacing the TDD module with a conventional classifier head—specifically, a linear layer that outputs the token—while keeping all other components identical. Concretely, this corresponds to attaching a linear classification head to the feature extractor shown in Fig.~\ref{fig:overview}. The results in Tab.~\ref{tab:main} are striking: TDD consistently outperforms MDD by a clear margin on the classification task. Our ``ViT-Base (TDD)'' model achieves a top-1 accuracy of 82.8\%, substantially higher than its fair counterpart ``ViT-Base (standard classification).'' Even more remarkably, ``ViT-Base (TDD)'' surpasses ``ViT-Large (standard classification)'' variants that employ nearly three times more parameters (305M vs.\ 111M). These comparisons strongly validate the effectiveness and efficiency of TDD.

\begin{figure}[t]
    \centering
    \includegraphics[width=0.9\linewidth]{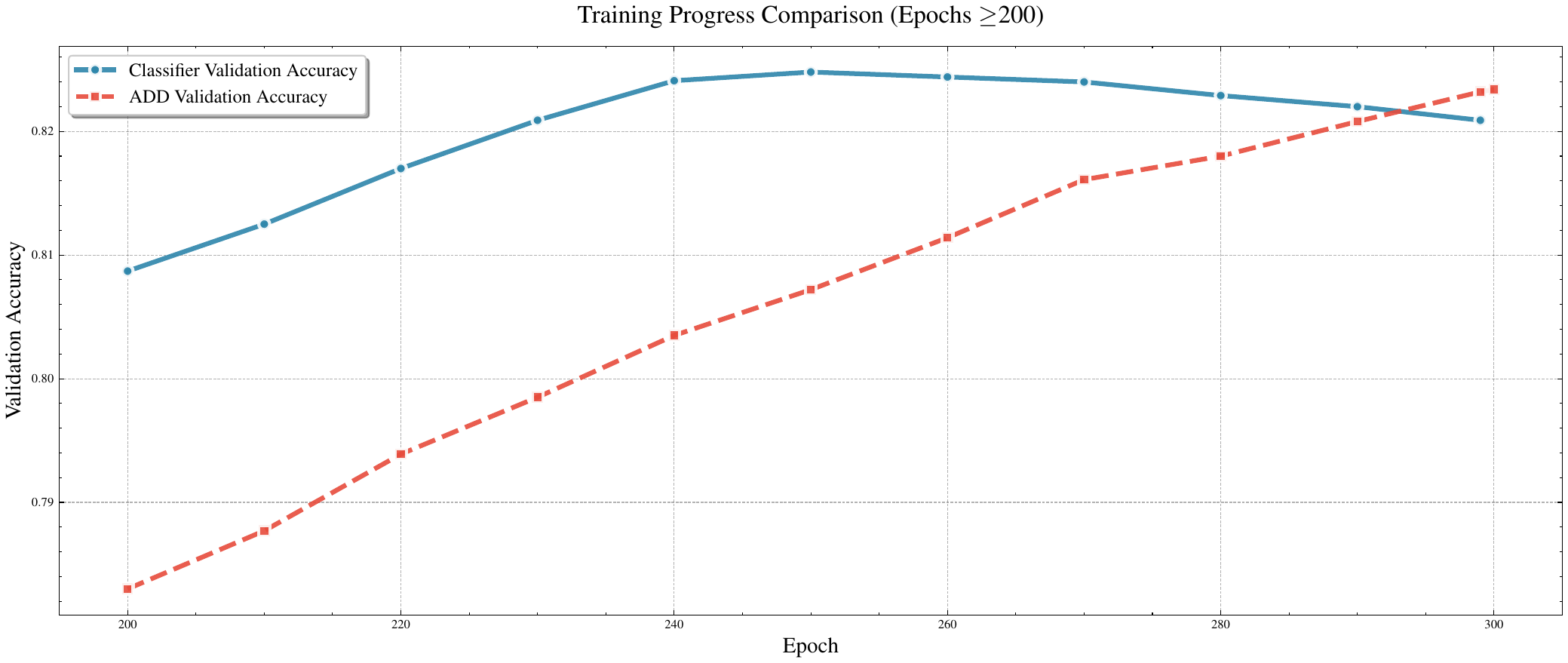}
    \caption{\textbf{Comparison of TDD and MDD Accuracy Across Epochs.} We trained TDD and our state-of-the-art method on ImageNet for 300 epochs, tracking accuracy at each epoch. The state-of-the-art method began to overfit after 250 epochs, whereas our method continued to improve beyond 300 epochs. This justifies extending the training to 500 epochs, ultimately achieving optimal performance (Top-1 score = 82.8).}
    \label{fig:training_curves}
\end{figure}

 To compare with standard methods, we implemented ViT-Base and successfully achieved an accuracy proposed in \citep{li2024autoregressive} of 82.3\%. We found that ViT-Base reached its highest accuracy around 250 epochs, after which it began to overfit and saw no further improvement. Our TDD method, however, reached the same accuracy as ViT-Base after 300 epochs and showed a trend of continued improvement. Therefore, we increased the number of TDD epochs until convergence. We finally achieved optimal performance of TDD after 500 epoch of training, 82.8\%. See Fig.~\ref{fig:training_curves} for a detailed comparison.

\begin{figure}[t]
    \centering
    \includegraphics[width=0.75\linewidth]{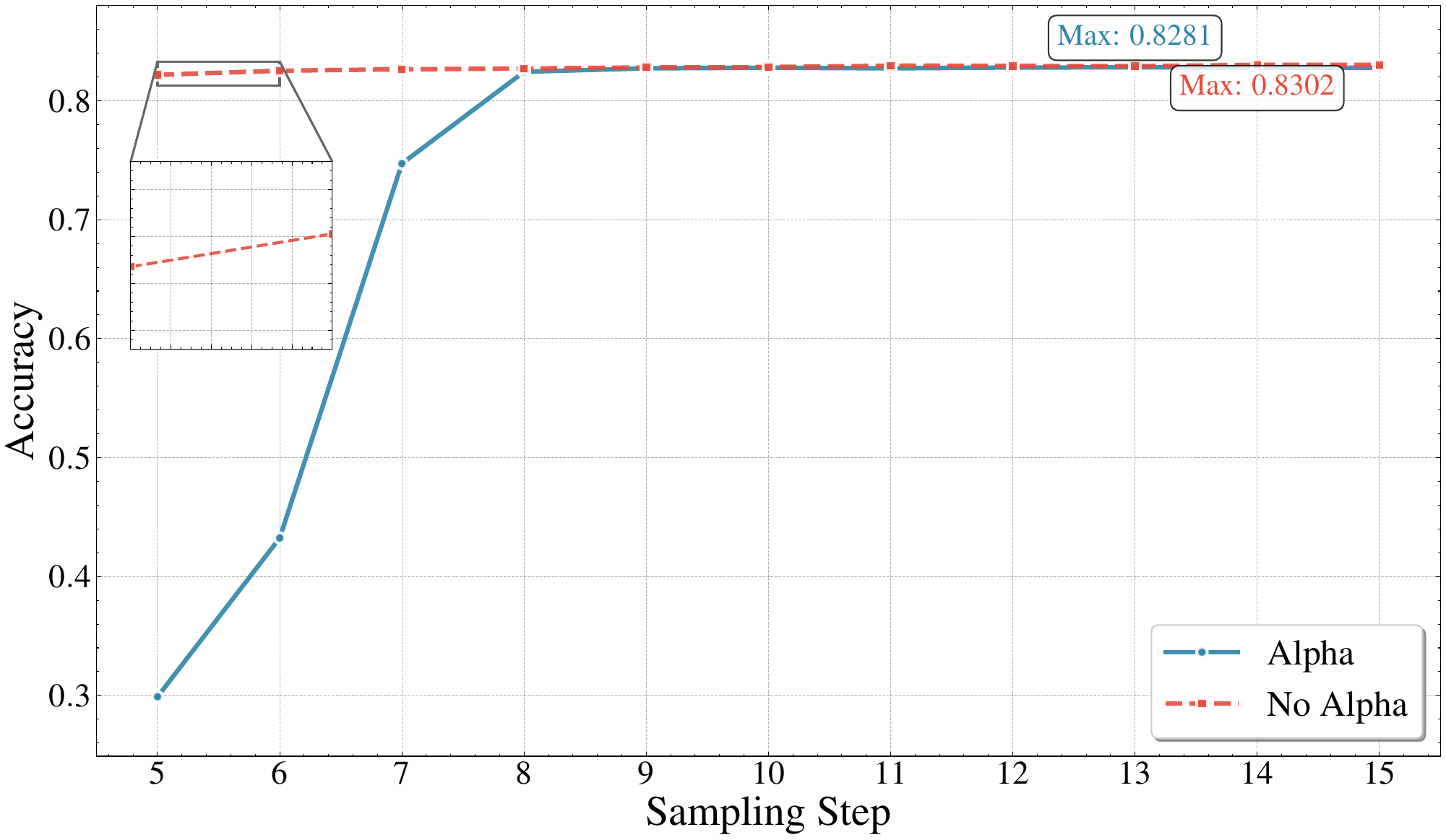}
    \caption{\textbf{Performance test of TDD at different sampling steps.} We conducted experiments with and without timestep-conditioned coefficients respectively, and tested the classification accuracy of TDD at different sampling steps. We found that TDD can achieve strong performance in about 10 sampling steps, and the best result is achieved after 20 iterations. However, without the timestep-conditioned coefficients, the iterative process exhibits slightly higher accuracy}
    \label{fig:sampling_steps}
\end{figure}

We also tested the classification accuracy of TDD on ImageNet at different sampling steps. We found that TDD can achieve good classification results in about 10 sampling steps, and the best result is achieved after 20 iterations. Details shown as Fig.~\ref{fig:sampling_steps}.

\begin{figure}[t]
    \centering
    \includegraphics[width=0.75\linewidth]{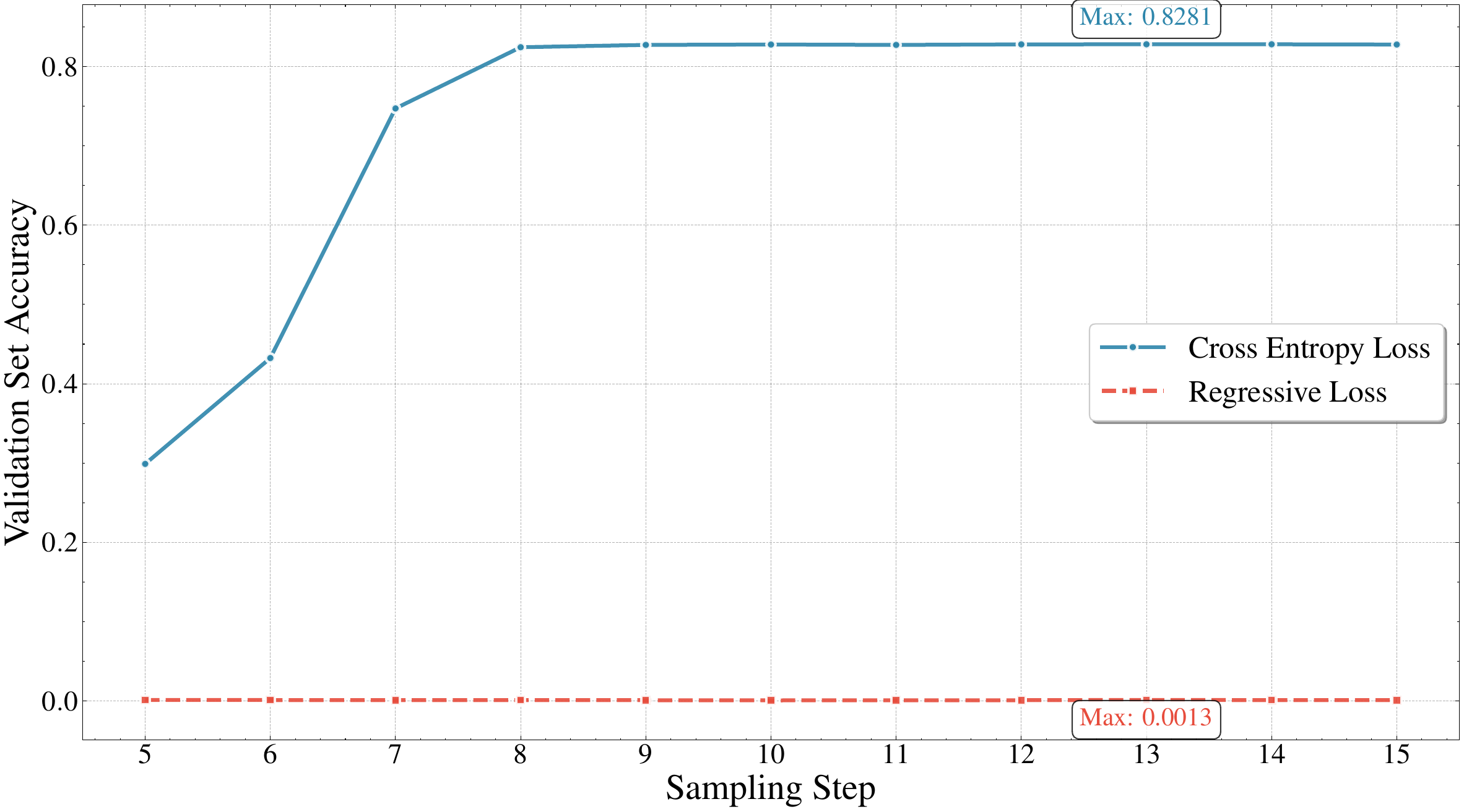}
    \caption{\textbf{Performance Comparison Between MSE Loss and Cross-Entropy Loss.} We compared the outputs generated by models trained with the two loss functions. Models trained with cross-entropy loss achieved high accuracy, while those trained with MSE loss showed unsatisfactory performance.}
    \label{fig:ce_mse}
\end{figure}

\subsubsection{Ablation Analysis}

We perform comprehensive ablations to understand the design choices behind our framework. All variants use the same Transformer backbone (ViT-Base~\citep{dosovitskiy2021image}, 111M parameters) and share identical training configurations. For controlled comparisons, each model is trained for a full 500 epochs to ensure convergence. These results show that incorporating several key techniques leads to the performance reported in Tab.~\ref{tab:main}.

\begin{table}[t]
\centering
\caption{\textbf{Ablation study.} 
We ablate different training and generation strategies for TDD. Timestep-conditioned cross-entropy loss, classifier-free guidance, and the ``$\arg\max$ + one-hot'' sampling scheme all lead to consistent improvements in Top-1 accuracy.}
\label{tab:ablation}
\resizebox{0.7\textwidth}{!}{%
\begin{tabular}{l|c|c}%
    \toprule
    \textbf{Methods} & Epochs & \textbf{Top-1 (\%)} \\
    \midrule

    TDD (w/ regression loss)
        & $400$ & $0.13$ \\
    \midrule

    TDD (w/ timestep-conditioned CE loss)
        & $400$ & $\mathbf{82.72}$ \\
    \midrule\midrule

    TDD (w/o classifier-free guidance)
        & $500$ & $82.36$ \\
    \midrule

    TDD (w/ classifier-free guidance)
        & $500$ & $\mathbf{82.82}$ \\
    \midrule\midrule

    TDD (w/ timestep-conditioned coefficients)
        & $500$ & $82.82$ \\
    \midrule

    TDD (w/o timestep-conditioned coefficients)
        & $500$ & $\mathbf{82.96}$ \\
    \midrule\midrule

    TDD (sampling with softmax)
        & $500$ & $82.35$ \\
    \midrule

    TDD (sampling with ``argmax + one-hot'')
        & $500$ & $\mathbf{82.82}$ \\
    \bottomrule

\end{tabular}
}
\end{table}

\paragraph{MSE Loss vs. Cross-Entropy Alignment Loss.} 
We first assess the necessity of timestep-conditioned cross-entropy alignment loss. In training diffusion models, regressive losses (e.g., MSE, $\mathcal{L}_1$) are typically used. In contrast, TDD employs timestep-conditioned cross-entropy to enforce prediction of $x_0$ (also denoted $x_\text{start}$). To test its importance, we replace timestep-conditioned cross-entropy with a regressive loss, training the model to predict noise as in conventional diffusion. Tab.~\ref{tab:ablation} and Fig. \ref{fig:ce_mse} show that this substitution causes catastrophic degradation (82.73\% $\rightarrow$ 0.13\%), confirming that timestep-conditioned cross-entropy is indispensable for effective discrete-space diffusion. A comparison of the iterative processes for the two loss functions can be found in Fig.~\ref{fig:ce_mse}.

\paragraph{The Timestep-Conditioned Coefficient in the Cross-Entropy Loss.} 
The effectiveness of timestep-conditioned coefficients in improving the iterative process is shown in Fig.~\ref{fig:sampling_steps}. With the timestep-dependent coefficient, the iterative process significantly enhances the model's ability to refine predictions, achieving high classification accuracy after 10 steps. In contrast, removing the coefficient results in slightly higher accuracy but more limited improvements during iteration.

\paragraph{Classifier-Free Guidance.}
Classifier-free guidance (CFG) is critical in continuous diffusion models for conditional generation. To test its relevance in our discrete setting, we removed CFG from TDD while keeping other components fixed. The performance dropped significantly (82.82\% $\rightarrow$ 82.36\%), as shown in Tab.~\ref{tab:ablation}, confirming that CFG is equally essential in discrete diffusion for ensuring semantic consistency with conditioning inputs. This result further supports our claim that TDD retains the core mechanics of standard diffusion models, affirming it as a \emph{Tweedie} Discrete Diffusion model.

\begin{figure}[t]
\vspace{-11pt}
    \centering
    \includegraphics[width=0.75\linewidth]{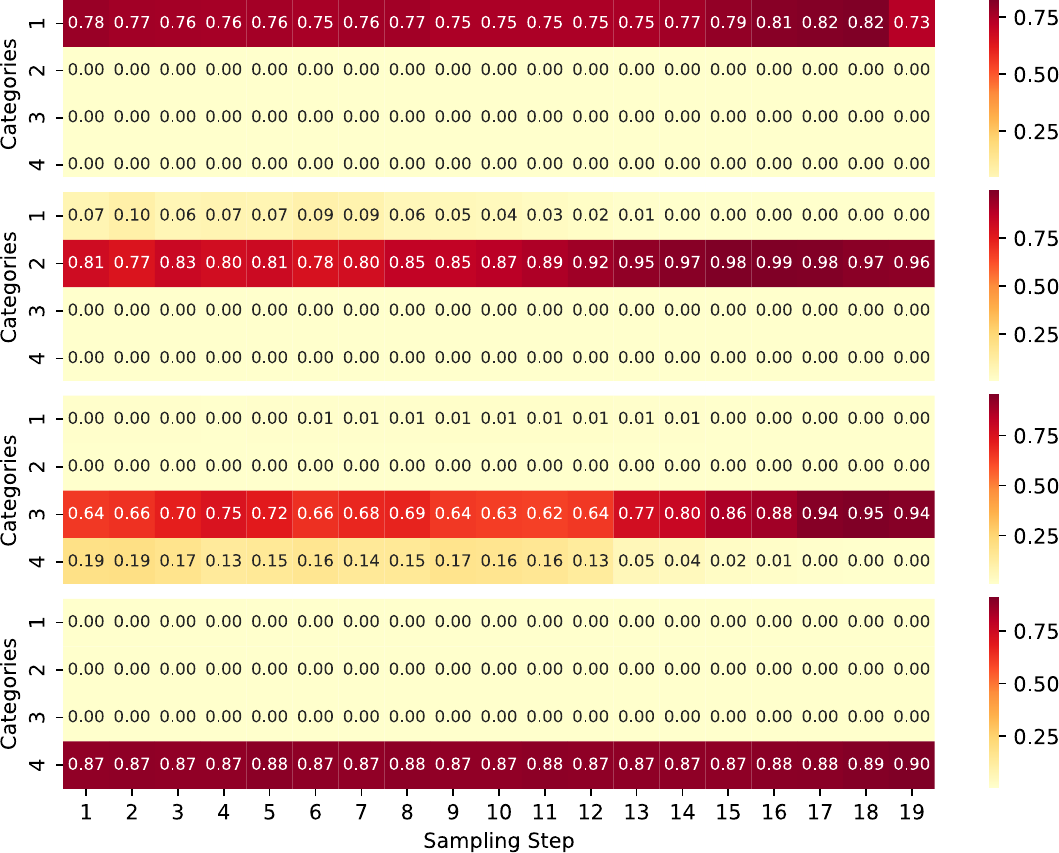}
    \caption{\textbf{The distribution change of model output with sampling steps.} To verify that the model can gradually generate one-hot labels through denoising, we visualized the output of the model at each sampling step. The results show that when the sampling step is set to 20, the model is able to gradually denoise pure Gaussian noise into an approximate one-hot vector.}
    \label{fig:heatmap}
\end{figure}

\paragraph{Sampling Strategy.}
Finally, we compare two ``to-one'' operations used during sampling: (i) softmax-based sampling and (ii) $\argmax$ followed by one-hot projection. As illustrated in Tab.~\ref{tab:ablation}, both strategies perform competitively, though $\argmax$ + one-hot projection yields a slight but consistent advantage. Consequently, unless otherwise specified, all sampling results in this paper use $\argmax$ + one-hot projection as the default strategy. For the specific accuracy changes resulting from different iteration settings during sampling, refer to Fig.~\ref{fig:sampling_steps}. Furthermore, Fig.~\ref{fig:heatmap} demonstrates that the output distribution of our model gradually approaches a one-hot distribution as the iterations proceed.

\subsection{Multi-Token Generation (Text Generation)}

\begin{figure}[t]
	\centering
	\includegraphics[width=0.75\linewidth]{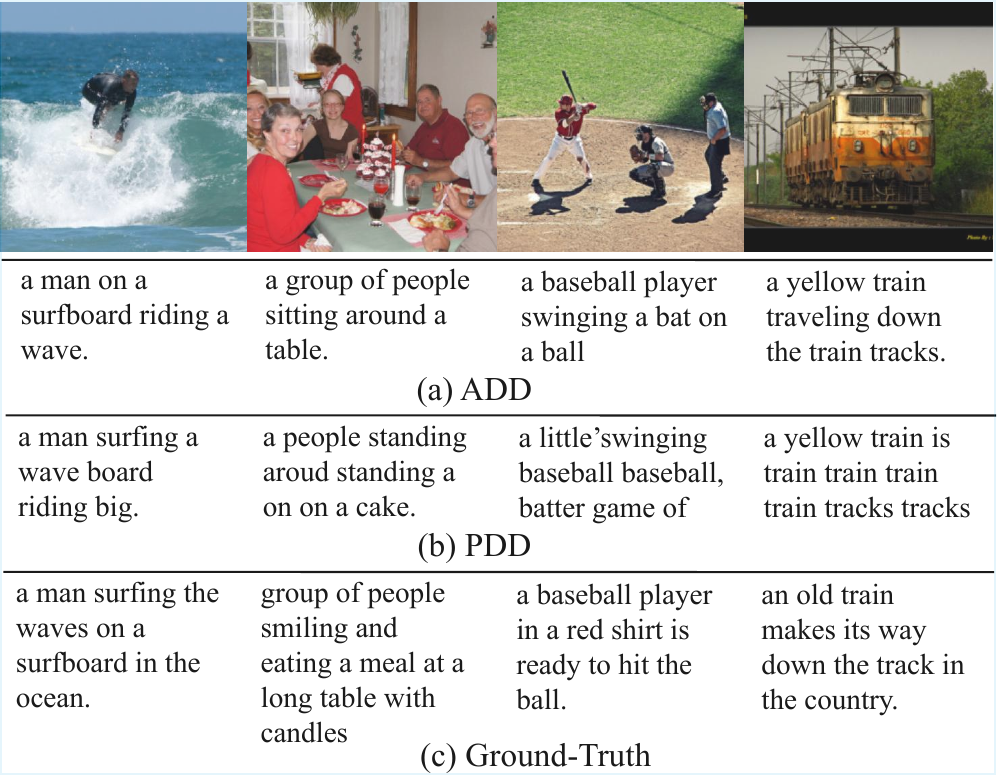}
	\caption{\textbf{Quality comparison of TDD and MDD.} The ground-truth captions are shown alongside captions generated by TDD and MDD for representative COCO examples. The superiority of our method over MDD is clearly demonstrated in the text comparison shown in the figure.}
	\label{fig:coco}
\end{figure}

\begin{table}[t]
\centering
\caption{\textbf{CLIP Scores metric testing and comparison.} CLIP Scores measure the cosine similarity between the normalized features of image-text pairs. In this part, we validated the similarity for three distinct types of such pairs.
}\label{tab:clipscore}
\resizebox{0.6\columnwidth}{!}{%
\begin{tabular}{l|c}
    \toprule
	\textbf{Selection of Text for Computation} & \textbf{CLIP Scores.} \\ \toprule
    Ground-Truth captions  & 0.30 \\ \midrule
    Shuffled captions      & 0.16 \\ \midrule
    MDD-Generated captions & 0.18 \\ \midrule
	\textbf{TDD-Generated captions} & 0.25 \\ \bottomrule
\end{tabular}%
}
\end{table}

We further investigate the potential of TDD in multi-token generation tasks by evaluating on ImageNet-1K-VL-Enriched and MS-COCO captioning datasets, with a particular focus on image captioning as a representative text generation benchmark. Unlike single-label classification, captioning requires the model to generate coherent sequences of tokens conditioned on visual inputs, thereby testing its ability to model discrete sequential dependencies. This task is especially challenging for diffusion-based approaches, as it requires capturing both the syntactic structure of natural language and the semantic alignment with images.

\begin{figure}[t]
	\centering
	\includegraphics[width=0.75\linewidth]{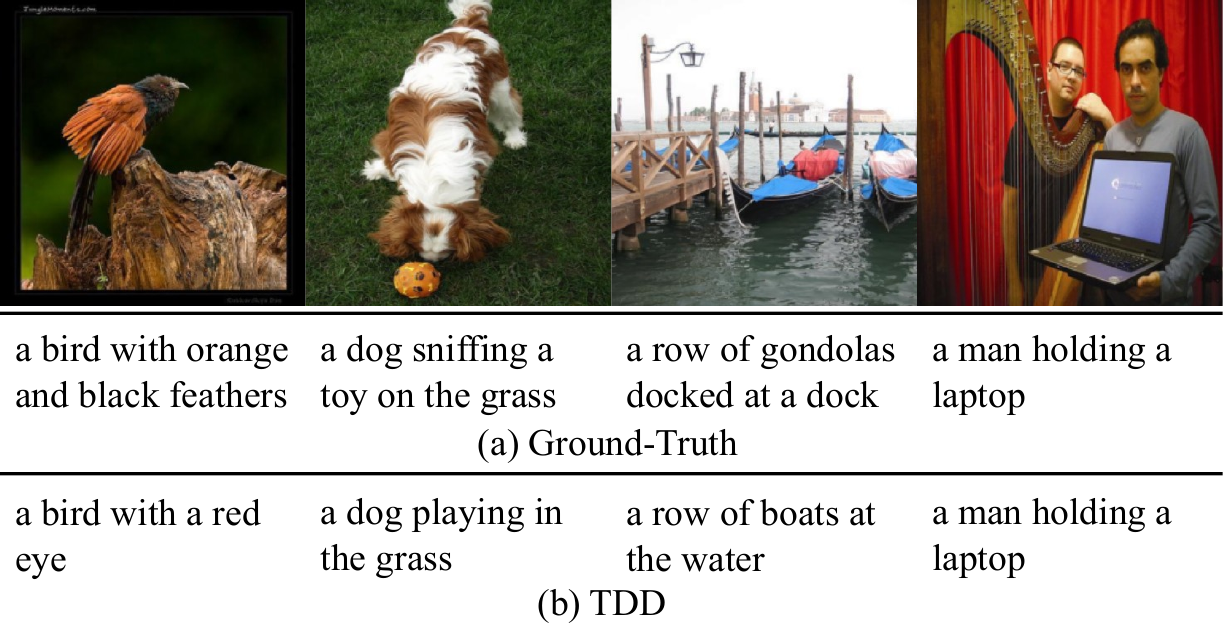}
	\caption{\textbf{Qualitative comparison of image captioning between ground-truth annotations and TDD-generated captions.}  TDD demonstrates strong semantic understanding across diverse visual scenes, accurately identifying objects and scenes.}
	\label{fig:imagenet-caption}
\end{figure}

We first present qualitative results on ImageNet-1K-VL-Enriched~\cite{imagenet-1k-vl-enriched}, which provides diverse image–text pairs for testing multi-token coherence. Representative captions on the validation split are shown in Fig.~\ref{fig:imagenet-caption}.

To mitigate the impact brought by the scarcity of text tokens, we utilized CLIP Scores as the evaluation metric. This metric employs CLIP to extract image and text features, and calculates the similarity between these two features, providing an well-suited evaluation of semantic relevance. In addition, we present qualitative examples to illustrate the fluency and semantic accuracy of generated texts.

For comparison, we benchmark our TDD against the masked diffusion framework, which we refer to as the masked discrete diffusion (MDD) model. MDD generates tokens by predicting masked entries in a partially observed sequence, but unlike TDD, it does not operate in a fully discrete diffusion process and therefore lacks consistent alignment with categorical token spaces.

To quantitatively assess the semantic relevance between captions and images, we employed a pre-trained CLIP model to extract features from both modalities and compute their similarity (Tab.~\ref{tab:clipscore}). We evaluated CLIP Scores for genuine image-text pairs, mismatched pairs, MDD-generated captions, and TDD-generated captions. The results clearly show that TDD-generated captions exhibit strong semantic alignment with the corresponding images.

Beyond numerical evaluation, qualitative comparisons in Fig.~\ref{fig:coco} show that TDD-generated captions are grammatically correct, semantically coherent, and well-aligned with both the input image and human-annotated ground-truth captions. In contrast, MDD outputs often exhibit broken syntax and poor semantic fidelity, with phrases disconnected from the visual content. These results support our claim that TDD, by operating directly in the one-hot label space, maintains both the discrete structure of text tokens and their semantic alignment with conditioning inputs.

Taken together, these results highlight two key insights. First, TDD substantially improves text generation quality over masked-discrete alternatives. Second, our findings suggest that TDD can serve as a foundation for broader applications in multi-token generation, bridging the gap between discrete modeling in vision tasks and natural language generation. This dual capability further strengthens the generality of TDD as a unified framework for discrete generative modeling.
\section{Conclusion}
\label{sec:conclusion}

We introduced \textbf{Tweedie Discrete Diffusion} (TDD), a diffusion framework for categorical data that leverages cross-entropy alignment, $\argmax$ discretization, and iterative denoising. Unlike masked-discrete methods, TDD denoises directly in one-hot space.
Experiments demonstrate that TDD delivers strong performance with very few sampling steps, surpassing masked-discrete baselines. Ablations further show that cross-entropy alignment and classifier-free guidance are essential for stable discrete diffusion.
Overall, TDD provides a principled route for applying diffusion models to symbolic domains, with promising extensions to large-scale language modeling.

\appendix

\section{Implementation Details}
\label{appendix:setup}
\subsection{Setup}
\paragraph{Datasets and Augmentation.} 
As mentioned in the experimental section, our classification experiments are conducted on the ImageNet\citep{5206848} dataset, incorporating a diverse set of advanced data augmentation strategies. 

For superior classification performance, we employ a comprehensive data augmentation pipeline. 
During training, images are randomly resized and cropped to the target resolution, followed by horizontal flipping. We use the AutoAugment policy \textit{rand-m9-mstd0.5-inc1}, which introduces a diverse set of photometric and geometric transformations. Color jitter is disabled in our default setting.

We further adopt Random Erasing with probability 0.25, using the \textit{pixel} mode and a single erase block. All images are normalized using the standard ImageNet mean and standard deviation. 

For regularization, we apply label smoothing with a smoothing factor of 0.1. In our primary experiments, Mixup and CutMix are disabled (mixup = 0, cutmix = 0). When enabled, mixing is performed with probability 1.0, switching to CutMix with probability 0.5, under a batch-level mixing strategy.

During evaluation, the input is resized using bicubic interpolation with a crop ratio, followed by a center crop to the target size. Finally, images are converted to tensors and normalized using the same ImageNet statistics as in training.

For text generation, TDD is trained on the ImageNet-1K-VL-Enriched\citep{imagenet-1k-vl-enriched} dataset together or the COCO\citep{lin2014microsoft} dataset.

For large-scale image–text pre-training in the captioning task, we adopt the ImageNet-1K-VL-Enriched dataset, which is built on the standard ImageNet-1K classification benchmark and further enriched with automatically generated captions, object-level bounding boxes and auxiliary metadata. Each sample comprises the original image, its ImageNet class label, a textual caption generated by BLIP2, and additional information that is not utilized in our experiments. This enriched corpus thereby transforms the original classification dataset into a versatile vision–language resource while retaining the taxonomy and visual diversity of ImageNet.

To complement ImageNet-1K-VL-Enriched with richly annotated everyday scenes and enable a more comprehensive and unbiased evaluation of TDD, we employ the COCO 2017 dataset. COCO (Common Objects in Context) consists of approximately 118 287 training images and 5 000 validation images. Each image is annotated with five human-written captions and object annotations across 80 common categories. In our experiments, we use the official caption annotations for training and validation. Owing to its emphasis on natural, multi-object and context-rich scenes, COCO 2017 provides an ideal benchmark for evaluating the semantic fidelity and linguistic diversity of our model.

\paragraph{Training Protocol.} 
TDD is trained from scratch for \textbf{500 epochs} distributed across $8 \times$ NVIDIA A100 80GB GPUs. All experiments are conducted using PyTorch~2.7.1 with CUDA~12.8. 

\paragraph{Evaluation Metrics.} 
For ImageNet classification, we report \textit{Top-1} accuracy on the validation set.  

For COCO captioning, our primary metric is \textit{CLIP Scores} computed between image and text features; higher is better.

\subsection{Samping Design}
During sampling, we designed two distinct approaches for processing the model's output at each iteration, as detailed below. 

The first is argmax sampling, which directly selects the class with the highest probability from the output and projects it into a one-hot space. This can be viewed as a hard probability space. 

The second approach is multinomial sampling, which first transforms the model's output into a soft probability space via softmax, then employs the torch.multinomial function for probability-based sampling, and finally maps the result to a cleaner one-hot space.

\subsection{Classifier-Free Guidance and Exponential Moving Average}

Classifier-Free Guidance (CFG) and Exponential Moving Average (EMA) are two essential components in our generation pipeline. CFG enables controllable conditioning strength during sampling, while EMA stabilizes model parameters to improve sample quality and evaluation robustness. We describe both mechanisms and detail how they are implemented in our framework.

\paragraph{Classifier-Free Guidance.}
CFG is a widely adopted technique in conditional generative modeling. Given a conditional prediction $x_{\text{cond}}$ and an unconditional prediction $x_{\text{uncond}}$, the final guided prediction is computed as:
\begin{equation}
    x_{\text{guided}}
    = x_{\text{uncond}} + \text{cfg} \cdot (x_{\text{cond}} - x_{\text{uncond}}),
\end{equation}
where $\text{cfg} \ge 1$ controls the guidance strength. A value of $\text{cfg}=1$ corresponds to unguided sampling, while larger values amplify semantic alignment with conditioning information.

In our implementation, CFG is applied at every autoregressive step during token sampling. The model learns a dedicated unconditional class embedding, allowing conditional and unconditional forward passes to be computed jointly.

\begin{algorithm}[t]
\caption{Sampling with CFG}
\KwIn{Model $f$, initial latent $x$, guidance scale $\text{cfg}$}
\For{$t=1$ \KwTo $T$}{
    Compute $u_t = f(x_t, \text{null})$\;
    Compute $c_t = f(x_t, \text{cond})$\;
    $\gamma_t \leftarrow s(t)$\;
    $x_t = u_t + \gamma_t(c_t - u_t)$\;
    Sample token and update $x_{t+1}$\;
}
\end{algorithm}

This sampling design enables TDD to balance semantic fidelity and sample diversity while adapting guidance intensity throughout the iterative process.

\paragraph{Exponential Moving Average.}
EMA maintains a smoothed copy of model parameters, which is updated after every training step:
\begin{equation}
    \theta_{\text{EMA}} \leftarrow \alpha\, \theta_{\text{EMA}}
    + (1 - \alpha)\, \theta,
\end{equation}
where $\alpha$ is typically close to~1 (we use $\alpha=0.9999$). This moving average reduces the stochastic noise of parameter updates and produces more stable convergence.

In practice, our system keeps two parameter sets: the raw parameters $\theta$ and the smoothed EMA parameters $\theta_{\text{EMA}}$. During evaluation, we switch to the EMA parameters, as they consistently yield better captioning quality and reduced variance across runs.

\begin{algorithm}[t]
\caption{EMA Update During Training}
\label{alg:ema_update}
\KwIn{Model parameters $\theta$, EMA parameters $\theta_{\text{EMA}}$, decay rate $\alpha$}
\For{each training iteration}{
    Compute gradients and update $\theta$\;
    \For{each parameter $\theta_i$}{
        $\theta_{\text{EMA}, i} \leftarrow 
        \alpha \cdot \theta_{\text{EMA}, i} + (1-\alpha)\cdot\theta_i$\;
    }
}
\KwRet{$\theta_{\text{EMA}}$}
\end{algorithm}

EMA therefore serves two purposes: (1) mitigating overfitting to noisy updates during training, and (2) ensuring that sampling and evaluation rely on a more stable, lower-variance parameter trajectory.

\section{Additional Experiments}

\subsection{Extended Analysis of TDD's Reverse Denoising Dynamics}
\label{appendix:tdd_extended}

\begin{figure*}[t]
    \centering
    \includegraphics[width=\textwidth]{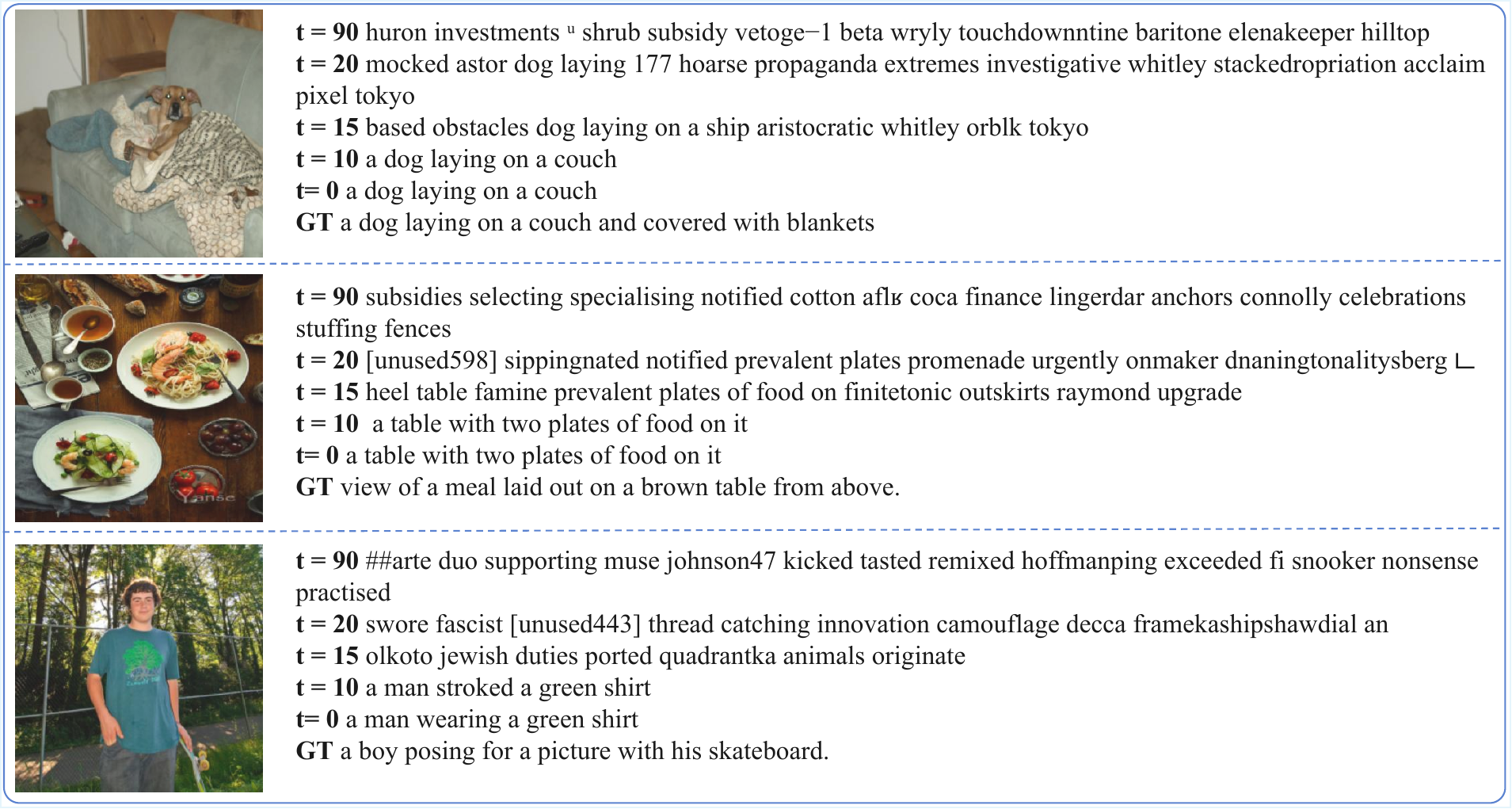}
    \caption{\textbf{Additional qualitative examples for TDD.} For each input image (left), we show TDD's textual predictions at multiple diffusion timesteps (right), ranging from highly corrupted discrete token states at large timesteps to coherent natural-language descriptions at $t=0$. The visualization reveals how semantic structure emerges gradually throughout the reverse denoising trajectory.}
    \label{fig:text-diffusion-trajectory}
\end{figure*}

In this appendix, we provide an extended qualitative analysis of the reverse denoising behavior exhibited by our TDD (Tweedie Discrete Diffusion) framework. The examples in Figure~\ref{fig:text-diffusion-trajectory} show how TDD transforms heavily corrupted discrete token states into coherent natural-language captions as the diffusion process proceeds toward lower timesteps. These visualizations expose the full semantic refinement path taken by the model, offering interpretability insights not observable from final-step predictions alone.

At large timesteps (e.g., $t = 90$), the model operates under high levels of corruption, producing token sequences with extremely high entropy. Because TDD injects Gaussian noise into a relaxed embedding representation before projecting back to the discrete one-hot space, the decoded outputs contain fragmented or fabricated lexical elements, exhibiting no meaningful alignment with the visual input. These states reflect the structural priors of the tokenizer rather than any grounded semantics.

As the reverse diffusion trajectory advances to intermediate timesteps (around $t = 20$--$15$), semantically relevant anchors begin to emerge. These anchors typically manifest as nouns corresponding to visually salient objects, but remain intermixed with substantial residual noise. This stage reveals the earliest point at which TDD begins to integrate visual cues into the discrete denoising pathway. Notably, object-centric tokens tend to surface before relational or structural linguistic elements, reflecting their greater robustness to noise.

In the later stages of the process ($t = 10 \rightarrow 0$), TDD consolidates the partially formed semantic structure into coherent sentences. Once key object terms stabilize, the model progressively recovers syntactic elements, attribute modifiers, and connective words, ultimately producing clean and grammatically valid captions. The discrete projection step—enforced at every iteration—encourages tokens to remain stable once they have emerged, reducing semantic blending and ensuring sharp, unambiguous outputs at $t=0$.

The examples also highlight several characteristic failure modes that complement the quantitative findings in the main paper. TDD sometimes captures the dominant object while overlooking secondary entities (e.g., detecting a person but not a skateboard), suggesting that finer-grained cues may be underrepresented in the visual encoder. Attribute-level distinctions—such as differentiating ``boy'' from ``man''—may also be lost under moderate noise, leading to subtle semantic drift during intermediate timesteps. Additionally, temporary deviations into irrelevant token clusters occasionally appear mid-trajectory but typically correct themselves as denoising proceeds.

Overall, these qualitative observations emphasize the interpretability advantages of discrete diffusion. TDD provides direct access to intermediate linguistic states, enabling detailed inspection of how semantic information accumulates and stabilizes across timesteps. The analysis here demonstrates a consistent progression from high-entropy noise to structured language and identifies promising directions for improving fine-grained semantic fidelity in future iterations of the model.


\subsection{Token-Level Denoising Behavior in TDD}
\label{appendix:tdd_token_trajectories}

\begin{figure}[tbhp]
    \centering
    \includegraphics[width=\textwidth]{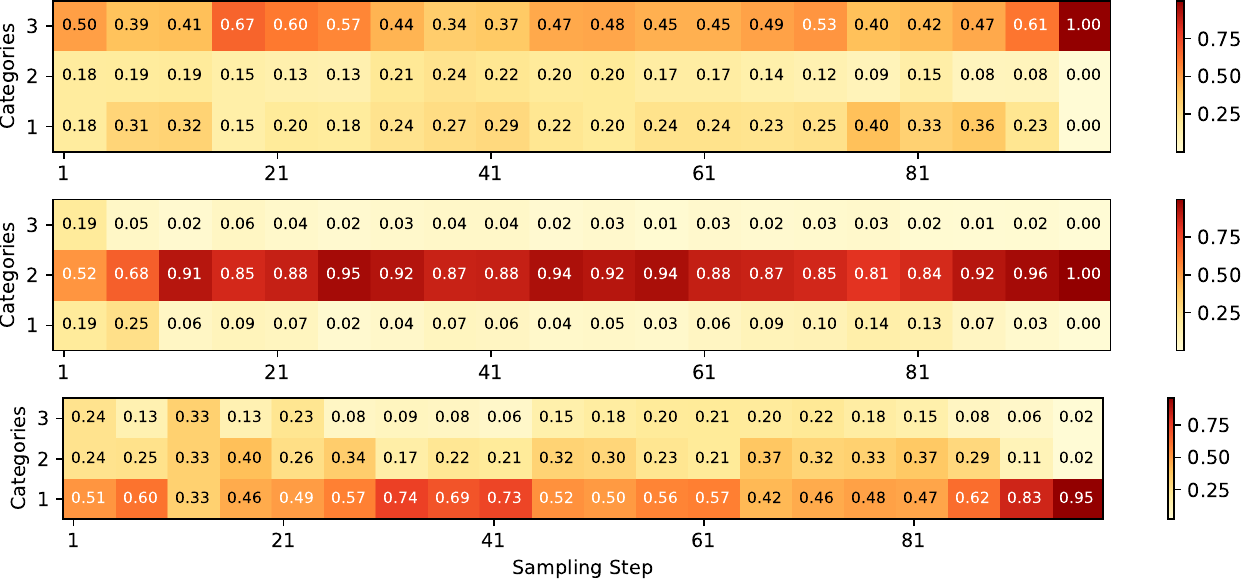}
    \caption{\textbf{Token-level trajectories during TDD denoising.} Each heatmap corresponds to a single token from the earlier qualitative examples and visualizes the categorical score assigned to several competing token classes across the diffusion sampling steps. These trajectories reveal how individual token semantics transition from high-entropy noise at large timesteps to stable linguistic interpretations as the model approaches $t=0$.}
    \label{fig:captioning-heatmap}
\end{figure}

In Figure~\ref{fig:text-diffusion-trajectory}, we illustrated the sentence-level denoising behavior of our TDD framework. To further analyze how individual tokens evolve during the reverse diffusion process, we provide in Figure~\ref{fig:captioning-heatmap} a set of heatmaps that visualize the categorical score trajectories of selected tokens across the full sampling path. Each heatmap corresponds to a single token taken from the earlier qualitative examples and plots the model's predicted scores for several candidate token categories as a function of the diffusion timestep. These trajectories reveal fine-grained semantic dynamics that are not captured by sentence-level visualizations alone.

At large timesteps, the model operates under heavy corruption, producing token-level predictions with high entropy. For some tokens, the model assigns nearly all probability mass to a dominant but incorrect category, showing that early reverse diffusion steps are largely governed by noise and embedding priors rather than grounded semantics. As denoising proceeds, the probability gradually shifts toward semantically meaningful categories, revealing the incremental emergence of linguistic structure.

Across the three heatmaps, we observe distinct patterns of token behavior. The first heatmap illustrates a token undergoing clear semantic correction: its dominant category transitions from an incorrect class at early timesteps toward the correct class as visual and contextual evidence accumulates. This behavior corresponds to tokens in the sentence-level visualizations that transform from incoherent fragments into meaningful words. The second heatmap depicts a token whose prediction remains stable throughout the entire trajectory. Such tokens are typically function words or high-confidence content words that are easily recoverable even under heavy noise, resulting in minimal fluctuation. The third heatmap shows a token experiencing strong temporal competition between multiple candidate categories. The model alternates among several plausible semantic hypotheses during intermediate steps before eventually converging to a final class. This reflects tokens in the sentence-level examples that exhibit ambiguity—often involving fine-grained distinctions such as age, object type, or subtle attributes—before stabilizing near the end of the sampling process.

These token-level trajectories demonstrate the interpretability advantages of discrete diffusion. Because TDD performs explicit denoising in one-hot token space, the intermediate states remain directly interpretable as categorical distributions, enabling us to observe how semantics are formed, refined, and occasionally revised as diffusion progresses. This analysis complements the sentence-level qualitative results and provides a deeper understanding of how discrete diffusion reconstructs linguistic structure from noise, shedding light on both the strengths of TDD’s semantic refinement process and its characteristic failure modes, such as late-stage ambiguity resolution or transient class switching.

\bibliography{main}

\end{document}